\documentclass[journal]{IEEEtran}

\ifCLASSINFOpdf
  \usepackage[pdftex]{graphicx}
\else

\fi
\usepackage[pdftex]{graphicx}
\usepackage{hyperref}
\usepackage{caption}
\urldef\myurl\url{https://vcl.iti.gr/project/epikoinono/}
\urldef\dataseturl\url{https://zenodo.org/record/3941811#.XxrZXZZRU5k}
\usepackage{tabularx,booktabs}

\usepackage{soul}
\newcolumntype{C}{>{\centering\arraybackslash}X}

\usepackage{amsmath}
\usepackage{amssymb}

\IEEEoverridecommandlockouts
\usepackage{alphabeta}

\begin{document}
\bstctlcite{IEEEexample:BSTcontrol}
\title{A Comprehensive Study on Deep Learning-based Methods for Sign Language Recognition}

\author{Nikolas Adaloglou\textsuperscript{1}\textsuperscript{*}\thanks{\textsuperscript{*}authors contributed equally},
        Theocharis Chatzis\textsuperscript{1}\textsuperscript{*},
       Ilias Papastratis\textsuperscript{1}\textsuperscript{*},
       Andreas Stergioulas\textsuperscript{1}\textsuperscript{*},
       Georgios Th. Papadopoulos\textsuperscript{1}, Member, IEEE, Vassia Zacharopoulou\textsuperscript{2}, George J. Xydopoulos\textsuperscript{2}, Klimnis Atzakas\textsuperscript{2}, Dimitris Papazachariou\textsuperscript{2}, and Petros Daras\textsuperscript{1}, Senior Member, IEEE \\

       The Visual Computing Lab, Information Technologies Institute, Centre for Research
and Technology Hellas\textsuperscript{1}\\
              University of Patras\textsuperscript{2}

\thanks{}}

\markboth{}%
{Shell \MakeLowercase{\textit{et al.}}: Bare Demo of IEEEtran.cls for IEEE Journals}
%



\maketitle
\begin{abstract}
In this paper, a comparative experimental assessment of computer vision-based methods for sign language recognition is conducted. By implementing the most recent deep neural network methods in this field, a thorough evaluation on multiple publicly available datasets is performed. The aim of the present study is to provide insights on sign language recognition, focusing on mapping non-segmented video streams to glosses. For this task, two new sequence training criteria, known from the fields of speech and scene text recognition, are introduced. Furthermore, a plethora of pretraining schemes is thoroughly discussed. Finally, a new RGB+D dataset for the Greek sign language is created. To the best of our knowledge, this is the first sign language dataset where three annotation levels are provided (individual gloss, sentence and spoken language) for the same set of video captures.

\end{abstract}

\begin{IEEEkeywords}
Sign Language Recognition, Greek sign language, Deep neural networks, stimulated CTC, conditional entropy CTC.
\end{IEEEkeywords}

\ifCLASSOPTIONpeerreview
\begin{center} \bfseries EDICS Category: 3-BBND \end{center}
\fi

\IEEEpeerreviewmaketitle

\section{Introduction}
Spoken languages make use of the ``{vocal - auditory}" channel, as they are articulated with the mouth and perceived with the ear. All writing systems also derive from, or are representations of, spoken languages. Sign languages (SLs) are different as they make use of the ``{corporal - visual}" channel, produced with the body and perceived with the eyes. SLs are not international and they are widely used by the communities of the Deaf. They are natural languages since they are developed spontaneously wherever the Deaf have the opportunity to congregate and communicate mutually \cite{sandler2006sign}. SLs are not derived from spoken languages; they have their own independent vocabularies and their own grammatical structures \cite{sandler2006sign}. The signs used by the Deaf, actually have internal structure in the same way as spoken words. Just as hundreds of thousands of English words are produced using a small number of different sounds, the signs of SLs are produced using a finite number of gestural features. Thus, signs are not holistic gestures but are rather analyzable, as a combination of linguistically significant features. Similarly to spoken languages, SLs are composed of the following indivisible features:

\begin{itemize}
  \item Manual features, i.e. hand shape, position, movement, orientation of the palm or fingers, and
  \item Non-manual features, namely eye gaze, head-nods/ shakes, shoulder orientations, various kinds of facial expression as mouthing and mouth gestures.
\end{itemize}

Combinations of the above-mentioned features represent a gloss, which is the fundamental building block of a SL and represents the closest meaning of a sign \cite{yang2019sf}. SLs, similar to the spoken ones, include an inventory of flexible grammatical rules that govern both manual and non-manual features \cite{koller2015continuous}. Both of them, are simultaneously (and often with loose temporal structure) used by signers, in order to construct sentences in a SL. Depending on the context, a specific feature may be the most critical factor towards interpreting a gloss. It can modify the meaning of a verb, provide spatial/temporal reference and discriminate between objects and people.

Due to the intrinsic difficulty of the Deaf community to interact with the rest of the society (according to \cite{mitchell2006many}, around 500,000 people use the American SL to communicate in the USA), the development of robust tools for automatic SL recognition would greatly alleviate this communication gap. As stated in \cite{bragg2019sign}, there is an increased demand for interdisciplinary collaboration including the Deaf community and for the creation of representative public video datasets.

Sign Language Recognition (SLR) can be defined as the task of inferring glosses performed by a signer from video captures. Even though there is a significant amount of work in the field of SLR, a lack of a complete experimental study is profound. Moreover, most publications do not report results in all available datasets or share their code. Thus, experimental results in the field of SL are rarely reproducible and lacking interpretation.
Apart from the inherent difficulties related to human motion analysis (e.g. differences in the appearance of the subjects, the human silhouette features, the execution of the same actions, the presence of occlusions, etc.) \cite{papadopoulos2016human}, automatic SLR exhibits the following key additional challenges:

\begin{itemize}
  \item The Deaf often employ a grammatical device known as ``{role-shifting}" when narrating an event or a story with more that one characters \cite{padden1986verbs}. Therefore, exact position in surrounding space and context have a large impact on the interpretation of SL. For example, personal pronouns (e.g. ``he", ``{she}", etc.) do not exist. Instead, the signer points directly to any involved referent or, when reproducing the contents of a conversation, pronouns are modeled by twisting his/her shoulders or gaze. Additionally, the Deaf leverage the space in front of them (signing space) in order to localize people or places \cite{emmorey2001space}. The latter is referred as placement in most SLs. By placing a person or a city somewhere in his/her signing space, the singer can refer to a person by pointing in its assigned space or show where a place is located, relative to the placed city.
  
  \item Many glosses are only distinguishable by their constituent non-manual features and they are typically difficult to be accurately detected, since even very slight human movements may impose different grammatical or semantic interpretations depending on the context \cite{cooper2011sign}.
  \item The execution speed of a given gloss may indicate a different meaning or the particular signer’s attitude. For instance, signers would not use two glosses to express ``{run quickly}", but they would simply speed up the execution of the involved signs \cite{cooper2011sign}.
  \item Signers often discard a gloss sub-feature, depending on previously performed and proceeding glosses. Hence, different instances of the exact same gloss, originating even from the same signer, can be observed.
  \item For most SLs so far, very few formal standardization activities have been implemented, to the extent that signers of the same country exhibit distinguishable differences during the execution of a given gloss \cite{ronchetti2016lsa64}.
\end{itemize}

Historically, before the advent of deep learning methods, the focus was on identifying isolated glosses and gesture spotting. Developed methods were often making use of hand crafted techniques \cite{kadous1996machine}, \cite{wang2014superpixel}. For spatial representation of the different sub-gloss components, they usually used handcrafted features and/or fusion of multiple modalities. Temporal modeling was achieved by classical sequence learning models, such as Hidden Markov Model (HMM) \cite{evangelidis2014continuous}, \cite{zhang2016chinese}, \cite{koller2018deep} and hidden conditional random fields \cite{wang2006hidden}. The rise of deep networks was met with a significant boost in performance for many video-related tasks, like human action recognition \cite{donahue2015long}, \cite{feichtenhofer2016convolutional}, gesture recognition, \cite{molchanov2015hand}, \cite{camgoz2016using}, motion capturing \cite{alexiadis2016integrated}, \cite{alexiadis2014quaternionic}, etc. SLR is a task closely related to computer vision. This is the reason that most approaches tackling SLR have adjusted to this direction.

In this paper, SLR using Deep Neural Network (DNN) methods is investigated. The main contributions of this work are summarized as follows:

\begin{itemize}
  \item A comprehensive, holistic and in-depth analysis of multiple literature DNN-based SLR methods is performed, in order to provide meaningful and detailed insights to the task at hand.
  \item Two new sequence learning training criteria are proposed, known from the fields of speech and scene text recognition.
  \item A new pretraining scheme is discussed, where transfer learning is compared to initial pseudo-alignments.
   \item A new publicly available large-scale RGB+D Greek Sign Language (GSL) dataset is introduced, containing real-life conversations that may occur in different public services. This dataset is particularly suitable for DNN-based approaches that typically require large quantities of expert annotated data.
\end{itemize}

The remainder of this paper is organized as follows: in Section \ref{sec:related}, related work is described. In Section \ref{sec:datasets}, an overview of the publicly available datasets in SLR is provided, along with the introduction of a new GSL dataset. In Section \ref{sec:soaMethods}, a description of the implemented architectures is given. In Section \ref{sec:sequence-learning-training-criteria-SLR}, a description of the proposed sequence training criteria is detailed. In Section \ref{sec:Experimentals}, the performed experimental results are reported. Then, in Section \ref{sec:Discussion}, interpretations and insights of the conducted experiments are discussed. Finally, conclusions are drawn and future research directions are highlighted in Section \ref{sec:conclusion}.

\section{Related Work} \label{sec:related}

\begin{figure*}[!t]
 \centering
 \includegraphics[width=0.9\linewidth]{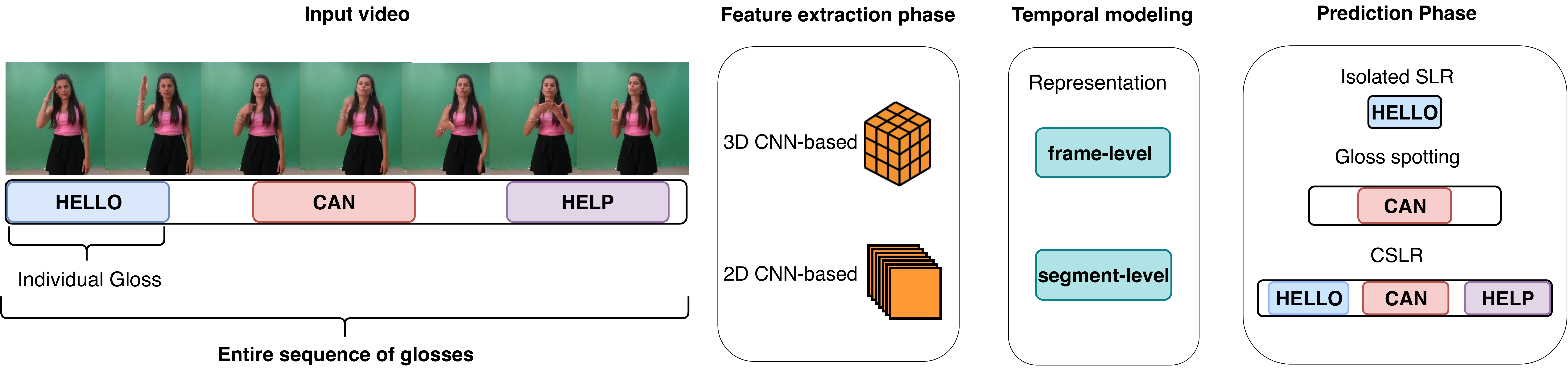}
  \captionsetup{justification=centering}
  \caption{An overview of SLR categories}
  \label{fig1}
\end{figure*}

The various automatic SLR tasks, depending on the modeling's level of detail and the subsequent recognition step, can be roughly divided in (\figurename{} \ref{fig1}):

\begin{itemize}
    \item Isolated SLR: Methods of this category target to address the task of video segment classification (where the segment boundaries are provided), based on the fundamental assumption that a single gloss is present \cite{konstantinidis2018deep}, \cite{cooper2012sign}, \cite{camgoz2016using}. 
    \item Sign detection in continuous streams: The aim of these approaches is to detect a set of predefined glosses in a continuous video stream \cite{evangelidis2014continuous}, \cite{neverova2015moddrop}, \cite{wu2016deep}. 
    \item Continuous SLR (CSLR): These methods aim at recognizing the sequence of glosses that are present in a continuous/non-segmented video sequence  \cite{koller2019weakly},  \cite{cui2019deep},  \cite{papastratis2020continuous}. This category of approaches exhibits characteristics that are most suitable for the needs of real-life SLR applications \cite{bragg2019sign}; hence, it has gained increased research attention and will be further discussed in the remainder of this section.
\end{itemize}

\subsection{Continuous sign language recognition}
By definition, CSLR is a task very similar to the one of continuous human action recognition, where a sequence of glosses (instead of actions) needs to be identified in a continuous stream of video data. However, glosses typically exhibit a significantly shorter duration than actions (i.e. they may only involve a very small number of frames), while transitions among them are often very subtle for their temporal boundaries to be efficiently recognized. Additionally, glosses may only involve very detailed and fine-grained human movements (e.g. finger signs or facial expressions), while human actions usually refer to more concrete and extensive human body actions. The latter facts highlight the particular challenges that are present in the CSLR field \cite{koller2015continuous}.

Due to the lack of gloss-level annotations, CSLR is regularly casted as a weakly supervised learning problem. The majority of CSLR architectures usually consists of a feature extractor, followed by a temporal modeling mechanism   \cite{camgoz2017subunets}, \cite{cui2017recurrent}. The feature extractor is used to compute feature representations from individual input frames (using 2D CNNs) or sets of neighbouring frames (using 3D CNNs). On the other hand, a critical aspect of the temporal modeling scheme enables the modeling of the SL unit feature representations (i.e., gloss-level, sentence-level). With respect to temporal modeling, sequence learning can be achieved using HMMs, Connectionist Temporal Classification (CTC) \cite{graves2006connectionist} or Dynamic Time Warping (DTW) \cite{sakoe1978dynamic} techniques. From the aforementioned categories, CTC has in general shown superior performance and the majority of works in CSLR has established CTC as the main sequence training criterion (for instance, HMMs may fail to efficiently model complex dynamic variations, due to expressiveness limitations \cite{cui2019deep}). However, CTC has the tendency to produce overconfident peak distributions, that are prone to overfitting \cite{liu2018connectionist}. Moreover, CTC introduces limited contribution towards optimizing the feature extractor \cite{zhou2019dynamic}. For these reasons, some recent approaches have adopted an iterative training optimization methodology. The latter essentially comprises a two-step process. In particular, a set of temporally-aligned pseudo-labels are initially estimated and used to guide the training of the feature extraction module. In the beginning, the pseudo-labels can be either estimated by statistical approaches \cite{koller2015continuous} or extracted from a shallower model \cite{cui2019deep}. After training the model in an isolated setup, the trained feature extractor is utilized for the continuous SLR setup. This process may be performed in an iterative way, similarly to the Expectation Maximization (EM) algorithm \cite{moon1996expectation}. Finally, CTC imposes a conditional independence constraint, where output predictions are independent, given the entire input sequence.

\subsection{2D CNN-based CSLR approaches}
One of the firstly deployed architectures in CSLR is based on \cite{koller2016deepHand}, where a CNN-HMM network is proposed. GoogleLeNet serves as the backbone architecture, fed with cropped hand regions and trained in an iterative manner. The same network architecture is deployed in a CSLR prediction setup \cite{koller2016deep}, where the CNN is trained using glosses as targets instead of hand shapes. Later on, in \cite{koller2017re}, the same authors extend their previous work by incorporating a Long Short-Term Memory unit (LSTM) \cite{hochreiter1997long} on top of the aforementioned network. In a more recent work \cite{koller2019weakly}, the authors present a three-stream CNN-LSTM-HMM network, using full frame, cropped dominant hand and signer's mouth region modalities. These models, since they employ HMM for sequence learning, have to make strong initial assumptions in order to overcome HMM's expressive limitations.

In \cite{camgoz2017subunets}, the authors introduce an end-to-end system in CSLR without iterative training. It consists of two streams, one responsible for processing the full frame sequences and one for processing only the signer's cropped dominant hand. In \cite{cui2017recurrent}, the authors employ a 2D CNN-LSTM architecture and in parallel with the LSTMs, a weakly supervised gloss-detection regularization network, consisting of stacked temporal 1D convolutions. The same authors in \cite{cui2019deep} extend their previous work by proposing a module composed of a series of temporal 1D CNNs followed by max pooling, between the feature extractor and the LSTM, while fully embracing the iterative optimization procedure. In \cite{yang2019sf}, a hybrid 2D-3D CNN architecture \cite{he2016deep} is developed. Features are extracted in a structured manner, where temporal dependencies are modeled by two LSTMs, without pretraining or using an iterative procedure. This approach however, yields the best results only in continuous SL datasets where a plethora of training data is available.

\subsection{3D CNN-based CSLR approaches}
One of the first works that employs 3D-CNNs in SLR is introduced in \cite{pu2016sign}. The authors present a multi-modal approach for the task of isolated SLR, using spatio-temporal Convolutional 3D networks (C3D) \cite{tran2015learning}, known from the research field of action recognition. Multi-modal representations are lately fused and fed to a Support Vector Machine (SVM) \cite{shawe2000support} classifier. The C3D architecture has also been utilized in CSLR by \cite{huang2018video}. The developed two-stream 3D CNN processes both full frame and cropped hand RGB images. The full network, named LS-HAN, consists of the proposed 3D CNN network, along with a hierarchical attention network, capable of latent space-based recognition modeling. In a later work \cite{joze2018ms}, the authors propose the I3D \cite{carreira2017quo} architecture in SLR. The model is deployed on an isolated SLR setup, with pretrained weights on action recognition datasets. The signer's body bounding box is served as input. For the evaluated dataset it yielded state-of-the-art results. In \cite{zhou2019dynamic}, the authors adopted and enhanced the original I3D model with a gated Recurrent Neural Network (RNN). Their aim is to accommodate features from different time scales.I3D has also been used as a baseline model in \cite{li2020word} on a large-scale isolated SLR dataset and achieved great recognition accuracy. In another work \cite{pu2018dilated}, the authors introduce the 3D-ResNet architecture to extract features. Furthermore, they substitute LSTM with stacked dilated temporal convolutions and CTC for sequence alignment and decoding. With this approach, they manage to have very large receptive fields while reducing time and space complexity, compared to LSTM. Finally, in \cite{pu2019iterative}, Pu \textit{et al.} propose a framework that also consists of a 3D-ResNet backbone. The features are provided in both an attentional encoder-decoder network \cite{bahdanau2014neural} and a CTC decoder for sequence learning. Both decoded outputs are jointly trained while the soft-DTW \cite{cuturi2017soft} is utilized to align them.

\begin{table*}[!t]
  \caption{Large-scale publicly available SLR datasets}
  \label{SLR-Datasets}
  \centering
    
\begin{tabularx}{\textwidth}{@{}l @{\extracolsep{\fill}} *{10}{c}@{}}
     \toprule
             & \multicolumn{10}{c}{\textbf{Characteristics}}\\
    \cmidrule(r){2-11}   
 
    \textbf{Datasets} & Language & Signers & Classes & Video instances & Duration (hours) & Resolution & fps & Type & Modalities & Year \\
    \midrule


Signum SI \cite{von2008significance}  & German & 25  & 780 & 19,500  & 55.3 & 776x578 & \textbf{30} & continuous & RGB & 2007 \\

Signum isol. \cite{von2008significance}  & German & 25 & 455 &  11,375  & 8.43 & 776x578 & \textbf{30} & both & RGB & 2007   \\

Signum subset \cite{von2008significance}  & German &  1  & 780 &  2,340  & 4.92 & 776x578 & \textbf{30} & both & RGB & 2007   \\

Phoenix SD \cite{forster2014extensions} & German &  9  & 1,231 &  6,841  & 10.71 & 210x260 & 25 & continuous & RGB & 2014 \\

Phoenix SI \cite{forster2014extensions} & German &  9  & 1,117 &  4,667  & 7.28 & 210x260 & 25 & continuous & RGB & 2014  \\

CSL SD \cite{huang2018video} & Chinese &  50  & 178 &  25,000 & \textbf{100+} & \textbf{1920x1080} & \textbf{30} & continuous & \textbf{RGB+D} & 2016  \\

CSL SI \cite{huang2018video} & Chinese &  50  & 178 &  25,000 & \textbf{100+} & \textbf{1920x1080} & \textbf{30} & continuous & \textbf{RGB+D} & 2016  \\

CSL isol. \cite{pu2016sign} & Chinese &  50  & 500 &  \textbf{125,000} & 67.75 & \textbf{1920x1080} & \textbf{30} & isolated & \textbf{RGB+D} & 2016  \\

Phoenix-T \cite{cihan2018neural} & German &  9  & \textbf{1,231} &  8,257 & 10.53 & 210x260 & 25 & continuous & RGB & 2018  \\

ASL 100 \cite{joze2018ms}  & English & 189  & 100 &  5,736 & 5.55 & varying & varying & isolated  & RGB & 2019 \\

ASL 1000 \cite{joze2018ms}  & English &  \textbf{222}  & 1,000 & 25,513 & 24.65 & varying & varying & isolated & RGB & 2019 \\

\hline
GSL isol. (new)    & Greek &   7  & 310 &  40,785  & 6.44 & 848x480 & \textbf{30} & isolated & \textbf{RGB+D} & 2019 \\
GSL SD (new) & Greek & 7 & 310 & 10,295  & 9.59  & 848x480 & \textbf{30} & continuous & \textbf{RGB+D} & 2019 \\
 GSL SI (new) & Greek &   7  & 310 &  10,295  & 9.59 & 848x480 & \textbf{30} & continuous & \textbf{RGB+D} & 2019 \\
    \bottomrule
  \end{tabularx} 
  \label{table:slr-datasets}
\end{table*}

\section{Publicly available datasets}  \label{sec:datasets}
Existing SLR datasets can be characterized as isolated or continuous, taking into account whether annotation are provided at the gloss (fine-grained) or the sentence (coarse-grained) levels. Additionally, they can be divided into Signer Dependent (SD) and Signer Independent (SI) ones, based on the defined evaluation scheme. In particular, in the SI datasets a signer cannot be present in both the training and the test set. In Table \ref{table:slr-datasets}, the following most widely known public SLR datasets, along with their main characteristics, are illustrated:

\begin{itemize}
    \item The Signum SI and the Signum subset \cite{von2008significance} include laboratory capturings of the German Sign Language. They are both created under strict laboratory settings with the most frequent everyday glosses.
    \item The Chinese Sign Language (CSL) SD, the CSL SI and the CSL isol.  datasets \cite{huang2018video} are also recorded in a predefined laboratory environment with Chinese SL words that are widely used in daily conversations. 
    \item The Phoenix SD \cite{forster2014extensions}, the Phoenix SI \cite{forster2014extensions} and the Phoenix-T \cite{cihan2018neural} datasets comprise videos of German SL, originating from the weather forecast domain.
    \item The American Sign Language (ASL) \cite{joze2018ms} dataset contains videos of various real-life settings. The collected videos exhibit large variations in background, image quality, lighting and positioning of the signers.
\end{itemize}

\subsection{The GSL dataset}  \label{sec:gsl}
\subsubsection{Dataset description} 
In order to boost scientific research in the deep learning era, large-scale public datasets need to be created. In this respect and with a particular focus on the case of the GSL recognition, a corresponding public dataset has been created in this work. In particular, a set of seven native GSL signers are involved in the capturings. The considered application includes cases of Deaf people interacting with different public services, namely police departments, hospitals and citizen service centers. For each application case, 5 individual and commonly met scenarios (of increasing duration and vocabulary complexity) are defined. The average length of each scenario is twenty sentences with the mean individual sentence length amounting to 4.23 glosses. Subsequently, each signer was asked to perform the pre-defined dialogues in GSL five consecutive times. In all cases, the simulation considers a Deaf person communicating with a single public service employee, while all interactions are performed in GSL (the involved signer performed the sequence of glosses of both agents in the discussion).

Overall, the resulting dataset includes 10,295 sentence instances, 40,785 gloss instances, 310 unique glosses (vocabulary size) and 331 unique sentences. For the definition of the dialogues in the identified application cases, the particularities of the GSL and the corresponding annotation guidelines, GSL linguistic experts are involved. 

The proposed Greek SLR dataset contains: a) temporal gloss annotations, b) sentence annotations, and c) translated annotations to the Modern Greek language. All the referenced annotations are performed in the same set of video captures. This is in contrast to other SL datasets that either contain only a small subset of isolated signs, or no translation to the spoken language. Thus, our dataset can serve as a benchmark for multiple SL tasks: isolated SLR, CSLR, and SL translation. This enables method evaluation from isolated to continuous SLR, or even SL translation, on the same videos.

\begin{figure}[t]
 \centering
 \includegraphics[width=\linewidth]{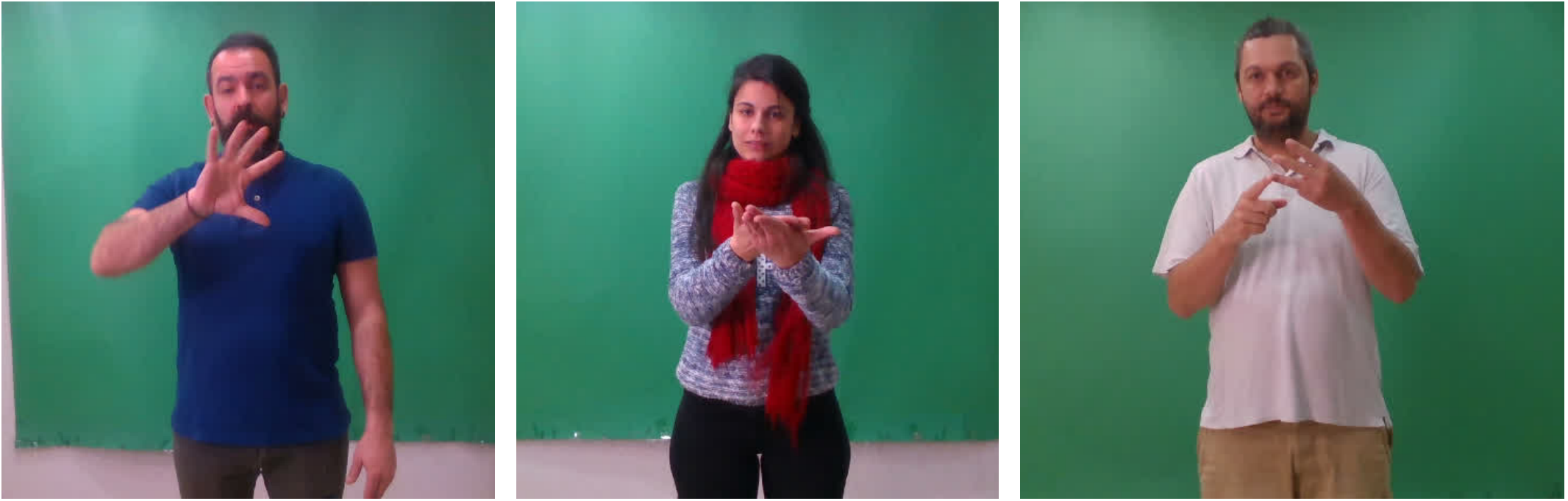}
  \caption{Example keyframes of the introduced GSL dataset}
  \label{fig:gsl-dataset}
\end{figure}

The recordings are conducted using an Intel RealSense D435 RGB+D camera at a rate of 30 fps. Both the RGB and the depth streams are acquired in the same spatial resolution of 848x480 pixels. To increase variability in videos, the camera position and orientation are slightly altered within subsequent recordings. Exemplary cropped frames of the captured videos are depicted in \figurename \ref{fig:gsl-dataset}.

\subsubsection{GSL evaluation sets}
Regarding the evaluation settings, the dataset includes the following setups: a) the continuous GSL SD, b) the continuous GSL SI, and c) the GSL isol. In GSL SD, roughly 80\% of the videos are used for training, corresponding to 8,189 instances. The rest 1,063 (10\%) are kept for validation and 1,043 (10\%) for testing. The selected test gloss sequences are not used in the training set, while all the individual glosses exist in the training set. In GSL SI, the recordings of one signer are left out for validation and testing (588 and 881 instances, respectively), which is approximately 14\% of the total data. The rest 8821 instances are utilized for training. A similar strategy is followed in GSL isol., where the validation set consists of 2,290 gloss instances, the test set 3,500, while the remaining 34,995 are used for training.

\subsubsection{Linguistic analysis and annotation process}
As already mentioned, the provided annotations are both at individual gloss and sentence level. Native signers annotated and labelled individual glosses, as well as whole sentences. Sign linguists and SL professional interpreters consistently validated the annotation of the individual glosses. A great effort was devoted in determining individual glosses following the ``{one form one meaning}" principle (i.e. a distinctive set of signs), taking into consideration the linguistic structure of the GSL and not its translation to the spoken standard modern Greek. We addressed and provided a solution for the following issues: a) compound words, b) synonyms, c) regional or stylistic variants of the same meaning, and d) agreement verbs.

In particular, compound words are composed of smaller meaningful units with distinctive form and meaning, i.e. the equivalent of morphemes of the spoken languages, which can also be simple individual words, for example: SON = MAN+BIRTH. Following the ``{one form one meaning}" principle, we split a compound word into its indivisible parts. Based on the above design, a computer vision system does not confuse compound words with its constituents.

Synonyms (e.g. two different signs with similar meaning) were distinguished to each other with the use of consecutively numbered lemmas. For instance, the two different signs which have the meaning: ``{DOWN}" were annotated as DOWN(1) and DOWN(2). The same strategy was opted for the annotation of regional and stylistic variants of the same meaning. For example, the two different regional variants of ``{DOCTOR} were annotated as DOCTOR(1), DOCTOR(2).

Another interesting case is the agreement verbs of sign languages, which contain the subject and/or object within the sign of the agreement verb. Agreement verbs indicate subjects and/or objects by changing the direction of the movement and/or the orientation of the hand. Therefore, it was decided that they cannot be distinguished as autonomous signs and are annotated as a single gloss. A representative example is the : ``{I DISCUSS WITH YOU}" versus ``{YOU DISCUSS WITH HIM}". For the described annotation guideline, the internationally accepted notation for the sign verbs is followed \cite{baker2016linguistics}, \cite{sandler2006sign}.

\section{SLR approaches}  \label{sec:soaMethods}
In order to gain a better insight on the behavior of the various automatic SLR approaches, the best performing and the most widely adopted methods of the literature are discussed in this section. The selected approaches cover all different categories of methods that have been proposed so far. The quantitative comparative evaluation of the latter will facilitate towards providing valuable insights for each SLR methodology.

\subsection{SubUNets}
Camgoz \textit{et. al} \cite{camgoz2017subunets} introduce a DNN-based approach for solving the simultaneous alignment and recognition problems, typically referred to as ``{sequence-to-sequence}" learning. In particular, the overall problem is decomposed of a series of specialized systems, termed SubUNets. Each SubUNet processes the frames of the video independently. Their model follows a 2D CNN-LSTM architecture, replacing HMM with LSTM-CTC. The overall goal is to model the spatio-temporal relationships among these SubUNets to solve the task at hand. More specifically, SubUNets allow to inject domain-specific expert knowledge into the system regarding suitable intermediate representations. Additionally, they also allow to implicitly perform transfer learning between different interrelated tasks. 


\subsection{GoogLeNet + TConvs}
In contrast to other 2D CNN-based methods that employ HMMs, Cui \textit{et. al} \cite{cui2019deep} propose a model that includes an extra temporal module (TConvs), after the feature extractor (GoogLeNet). The TConvs module consists of two 1D CNN layers and two max pooling layers. It is designed to capture the fine-grained dependencies, which exist inside a gloss (intra-gloss dependencies) between consecutive frames, into compact per-window feature vectors. The intermediate segment representations approximate the average duration of a gloss. Finally, bidirectional RNNs are applied in order to capture the context information between gloss segments. The total architecture is trained iteratively, in order to exploit the expressive capability of DNN models with limited data.

\subsection{I3D+BLSTM}
Inflated 3D ConvNet (I3D) \cite{carreira2017quo} was originally developed for the task of human action recognition. Compared to 2D CNNs, 3D CNNs are able to directly learn spatiotemporal features from frame sequences. As such, its application has demonstrated outstanding performance on isolated SLR \cite{joze2018ms,li2020word}. In particular, the I3D architecture is an extended version of GoogLeNet, which contains several 3D convolutional layers followed by 3D max-pooling layers. The key insight of this architecture is the endowing of the 2D sub-modules (filters and pooling kernels) with an additional temporal dimension. The time dimension depends mostly on frame rate. For this reason, the stride and pooling size in are designed to be asymmetric to the spatial dimensions. This methodology makes feasible to learn spatio-temporal features from videos, while it leverages efficient known architecture designs and parameters. In order to bring this model in CSLR setup, the spatio-temporal feature sequence is processed by an BLSTM, modeling the long-term temporal correlations. The whole architecture is trained iteratively with a dynamic pseudo-label decoding method.

\subsection{3D-ResNet+BLSTM}
Pu \textit{et al.} \cite{pu2019iterative} propose a framework that consists of a lightweight 3D CNN backbone (3D-ResNet-18 \cite{hara2018can}) with residual connections for feature extraction, as well as a BLSTM for sequence learning. Two different decoding strategies are performed, one with the CTC criterion and the other with an attentional decoder RNN. The glosses predicted by the attentional decoder are utilised to draw a warping path using a soft-DTW \cite{cuturi2017soft} alignment constraint. The warping paths display the alignments between glosses and video segments. The produced pseudo-alignments of the soft-DTW are then employed for iterative optimization.


\section{Sequence learning training criteria for CSLR} \label{sec:sequence-learning-training-criteria-SLR}
A summary of the notations used in this paper, is provided in this section, so as to enhance its readability and understanding. Let us denote by $U$ the label (i.e. gloss) vocabulary and by $blank$ the new blank token, representing the silence or transition between two consecutive labels. The extended vocabulary can be defined as $V=U \cup \{blank\} \in R^{L}$, where $L$ is the total number of labels. From now on, given a sequence $\boldsymbol{f}$ of length $F$, we denote its first and last $p$ elements by $\boldsymbol{f}_{1:p}$ and $\boldsymbol{f}_{p:F}$, respectively. An input frame sequence of length $N$ can be defined as $ \boldsymbol{X}= ( \boldsymbol{x}_{1}, .., \boldsymbol{x}_{N})$. The corresponding target sequence of labels (i.e. glosses) of length $K$ is defined as $\boldsymbol{y}= (y_{1}, .., y_{K}) $. In addition, let $\boldsymbol{G}_v=(\boldsymbol{g}_{v}^{1}, .., \boldsymbol{g}_{v}^{T}) \in R^{L \times T}$ be the predicted output sequence of a softmax classifier, where $T \leq N$ and $v \in V$. $\boldsymbol{g}_{v}^{t}$ can be interpreted as the probability of observing label $v$ at time-step $t$. Hence, $\boldsymbol{G}_v$ defines a distribution over the set $V^{T}\in R^{L\times T}$:

\begin{equation}
    p( \pi | \boldsymbol{X}) = \prod_{t=1}^{T} g_{\pi_{t}}^{t}, \forall \pi \in V^{T}
    \label{eq:p_distribution}
\end{equation}

The elements of $V^{T}$ are referred as paths and denoted by $\pi$. In order to map $\boldsymbol{y}$ to $\pi$, one can define a mapping function $B: V^{T} \mapsto U^{\leq{T}}$, with $U^{\leq{T}}$ being the set of possible labellings. $B$ removes repeated labels and blanks from a given path. Similarly, one can denote the inverse operation of $B$ as $B^{-1}$, that maps target labels to all the valid paths. From this perspective, the conditional probability of $\boldsymbol{y}$ is computed as:

\begin{equation}
    p( \boldsymbol{y} | \boldsymbol{X}) =\sum_{\pi \in B^{-1}(\boldsymbol{y})}  p( \pi | \boldsymbol{X})
    \label{eq:prob_paths}
\end{equation}

\subsection{Traditional CTC criterion} \label{sec:CTC}
Connectionist Temporal Classification (CTC) \cite{graves2006connectionist} is widely utilized for labelling unsegmented sequences. The time complexity of \eqref{eq:prob_paths} is $O(L^N K)$, which means that the amount of valid paths grows exponentially with $N$. To efficiently calculate $p(\boldsymbol{y}|\boldsymbol{X})$, a recursive formula is derived, which exploits the existence of common sub-paths. Furthermore, to allow for blanks in the paths, a modified gloss sequence $\boldsymbol{y}'$ of length $K'=2K + 1$ is used, by adding blanks before and after each gloss in $\boldsymbol{y}$. Forward and backward probabilities $\alpha_{t}(s)$ of $\boldsymbol{y}'_{1:s}$  at $t$ and $\beta_{t}(s)$ of $\boldsymbol{y}'_{s:K'}$ at $t$ are defined as:

\begin{equation}
    \alpha_{t}(s) \triangleq \sum_{B(\pi_{1:t})=\boldsymbol{y}'_{1:s}}\prod_{t'=1}^{t}g_{\pi_{t'}}^{t'}
    \label{eq:alphas}
\end{equation}

\begin{equation}
    \beta_{t}(s) \triangleq \sum_{B(\pi_{t:T})=\boldsymbol{y}'_{s:K' }}\prod_{t'=t}^{T}g_{\pi_{t'}}^{t'}
    \label{eq:betas}
\end{equation}

Therefore, to calculate $p(\boldsymbol{y}|\boldsymbol{X})$ for any $t$, we sum over all $s$ in $\boldsymbol{y}'$ as:
\begin{equation}
    p(\boldsymbol{y}|\boldsymbol{X})=\sum_{s=1}^{K'}\frac{\alpha_{t}(s)\beta_{t}(s)}{g_{\boldsymbol{y}'_{s}}^{t}}
\end{equation}

Finally, the CTC criterion is derived as:
\begin{equation}
    L_{ctc} = -\log{p(\boldsymbol{y}|\boldsymbol{X})}
    \label{eq:l_ctc}
\end{equation}

The error signal of $L_{ctc}$ with respect to $g_{v}^{t}$ is:

\begin{equation}
     \frac{\partial L_{ctc}}{\partial{g_{v}^{t}}}=-\frac{1}{p(\boldsymbol{y}|\boldsymbol{X})g_{v}^{t}}\sum_{\{\pi\in B^{-1}(\boldsymbol{y}),\pi_{t}=v\}} p(\pi|\boldsymbol{X})
     \label{eq:l_ctc_singal_error}
\end{equation}

From \eqref{eq:l_ctc_singal_error} it can be observed that the error signal is proportional to the fraction of all valid paths. As soon as a path dominates the rest, the error signal enforces all the probabilities to concentrate on a single path. Moreover, \eqref{eq:p_distribution} and \eqref{eq:l_ctc_singal_error} indicate that the probabilities of a gloss occurring at following time-steps are independent, which is known as the conditional independence assumption. For these reasons, two learning criteria are introduced in CSLR: a) one that encounters the ambiguous segmentation boundaries of adjacent glosses, and b) one that is able to model the intra-gloss dependencies, by incorporating a learnable language model during training (as opposed to other approaches that use it only during the CTC decoding stage).

\subsection{Entropy Regularization CTC}  
The CTC criterion can be extended \cite{liu2018connectionist} based on maximum conditional entropy \cite{jaynes1957information}, by adding an entropy regularization term $H$:

\begin{multline}
         H(p(\pi |\boldsymbol{y},\boldsymbol{X}))=-\sum_{\pi \in B^{-1}(\boldsymbol{y})}p(\pi |\boldsymbol{X},\boldsymbol{y})\log{p(\pi |\boldsymbol{X},\boldsymbol{y})}  
   \\=  -\frac{Q(\boldsymbol{y})}{p(\boldsymbol{y} |\boldsymbol{X})} + \log{p(\boldsymbol{y}|\boldsymbol{X}}),
    \label{eq:l_entropy_H}
\end{multline}

where $Q(\boldsymbol{y})=\sum_{\pi \in B^{-1}(\boldsymbol{y})} p(\pi | \boldsymbol{X})\log{p(\pi |\boldsymbol{X})}$.

\smallskip

$H$ aims to prevent the entropy of the non-dominant paths from decreasing rapidly. Consequently, the entropy regularization CTC criterion (EnCTC) is formulated as:

\begin{equation}
    L_{enctc} = L_{ctc} - \phi H(p(\pi | \boldsymbol{y}, \boldsymbol{X})),
    \label{eq:l_entc}
\end{equation}
where $\phi$ is a hyperparameter. The introduction of the entropy term $H$ prevents the error signal from gathering into the dominant path, but rather encourages the exploration of nearby ones. By increasing the probabilities of the alternative paths, the peaky distribution problem is alleviated.

\subsection{Stimulated CTC} \label{sec:StimCTC} 
Stimulated learning \cite{tan2015improving,wu2016stimulated,wu2017improving} augments the training process by regularizing the activations of the sequence learning RNN, $\boldsymbol{h}_{t}$. Stimulated CTC (StimCTC) \cite{heymann2019improving} constricts the independent assumption of traditional CTC. To generate the appropriate $stimuli$, an auxiliary uni-directional Language Model RNN (RNN-LM) is utilized. The RNN-LM encoded hidden states ($\boldsymbol{h}_{k}$) encapsulate the sentence's history, up to gloss $k$. $\boldsymbol{h}_{t}$ is stimulated by utilizing the non-blank forward and backward probabilities of the CTC criterion, denoted as $\boldsymbol{\alpha}'_{t}$ and $\boldsymbol{\beta}'_{t}$ $\in{R}^{K}$ respectively, and calculated as follows:

\begin{align}
    \boldsymbol{\alpha}'_{t,k} = \boldsymbol{\alpha}_{t,2k+1}\\
        \boldsymbol{\beta}'_{t,k} = \boldsymbol{\beta}_{t,2k+1}
\end{align}
Then, the weighting factor $\boldsymbol{\gamma}_{t}$ can be calculated as:

\begin{equation}
  \boldsymbol{\gamma}_{t}=\frac{\boldsymbol{\beta}'_{t}\odot\boldsymbol{\alpha}'_{t}}{\boldsymbol{\beta}'_{t}\cdot\boldsymbol{\alpha}'_{t}},
\end{equation}
where $\odot$ implies an element-wise multiplication.

Intuitively, $\boldsymbol{\gamma}_{t}$ can be seen as the probabilities of any gloss in target sequence $\boldsymbol{y}$ to be mapped to time-step $t$. The linguistic structure of SL is then incorporated as:
 \begin{equation}
    L_{stimuli} =\frac{1}{K\cdot T}\sum_{k}^{K}\sum_{t}^{T}\gamma_{t}(k){\mid\mid\boldsymbol{h}_{t}-\boldsymbol{h}_{k}\mid\mid}^2
\end{equation}  
Thereby, $\boldsymbol{h}_{t}$ is enforced to comply with $\boldsymbol{h}_{k}$. The RNN-LM model is trained using the cross-entropy criterion denoted as $L_{lm}$. Finally, the StimCTC criterion is defined as:

\begin{equation}
    L_{stim} = L_{ctc} + \lambda L_{lm} + \theta L_{stimuli},
\end{equation}
where $\lambda$ and $\theta$ are hyper-parameters. The described criteria can be combined, resulting in Entropy Stimulated CTC (EnStimCTC) criterion, as:
\begin{equation}
    L_{enstim} = L_{ctc} - \phi H(p(\pi | \boldsymbol{y})) + \lambda L_{lm} + \theta L_{stimuli}
\end{equation}

\section{Experimental evaluation} \label{sec:Experimentals}
In order to provide a fair evaluation, we re-implemented the selected approaches and evaluated them on multiple large-scale datasets, in both isolated and continuous SLR. Re-implementations are based on the original authors' guidelines and any modifications are explicitly referenced. For the continuous setup, the criteria CTC, EnCTC, and EnStimCTC are evaluated in all architectures. For a fair comparison between different models, we opt to use the RGB full frame modality, since it is the common modality between selected datasets and it is more suitable for real-life applications. In addition, we conduct experiments on GSL SI and SD datasets using the depth modality (Table \ref{table:depth}). 

We omit the iterative optimization process, instead we pretrain each model on the respective dataset's isolated version if present. Otherwise, extracted pseudo-alignments from other models (i.e. Phoenix) are used for isolated pretraining (implementations and experimental results are publicly available to enforce reproducibility in SLR\footnote[3]{\dataseturl}).

\subsection{Datasets and Evaluation metrics}
The following datasets have been chosen for experimental evaluation: ASL 100 and 1000, CSL isol., GSL isol. for the isolated setup, and Phoenix SD and Phoenix SI, CSL SD, CSL SI, GSL SD, GSL SI for the CSLR setup. To evaluate recognition performance in continuous datasets, the word error rate (WER) metric has been adopted, which quantifies the similarity between predicted glosses and ground truth gloss sequence. WER measures the least number of operations needed to transform the aligned predicted sequence to the ground truth and can be defined as: 
\begin{equation}
    WER = \frac{S + D + I}{N},
\end{equation}
where $S$ is the total number of substitutions, $D$ is the total number of deletions, $I$ is the total number of insertions and $N$ is the total number of glosses in the ground truth.

\subsection{Data augmentation and implementation details}
The same data preprocessing methods are used for all datasets. Each frame is normalized by the mean and standard deviation of the ImageNet dataset. To increase the variability of the training videos, the following data augmentation techniques are adopted. Frames are resized to 256X256 and cropped at a random position to 224X224. Random frame sampling is used up to $80\%$ of video length. Moreover, random jittering of the brightness, contrast, saturation and hue values of each frame is applied. 

The models are trained with Adam optimizer with initial learning rate $\lambda_{0}=10^{-4}$. The learning rate is reduced to $\lambda_{i}=10^{-5}$ when validation loss does not improve for 15 consecutive epochs. For isolated SLR experiments, the batch size is set to 2. Videos are rescaled to a fixed length that is equal to the average gloss length of each dataset. For CSLR experiments, videos are downsampled to maximum length of 250 frames, if necessary. The batch size is set to 1, due to GPU memory constraints. The experiments are conducted in a NVIDIA GeForce GTX-1080 Ti GPU with $12$ GB of memory and $32$ GB of RAM. All models, depending on the dataset, require 10 to 25 epochs to converge.

\begin{table}[b]
\caption{Gloss Test Accuracy in percentage - isolated SLR}
\centering
\begin{tabularx}{\linewidth}{@{}l @{\extracolsep{\fill}} *{4}{c}@{}}
\toprule

             & \multicolumn{4}{c}{\textbf{Datasets}}\\
    \cmidrule(l){2-5}   

\textbf{Method} & ASL 1000 & ASL 100 & CSL isol. & GSL isol.\\
\midrule

 GoogLeNet+TConvs \cite{cui2019deep} & - & 44.92 & 79.31 & 86.03  \\
  3D-ResNet+BLSTM \cite{pu2019iterative} & - & 50.48 & 89.91 &  86.23  \\
 I3D+BLSTM \cite{carreira2017quo}  & \textbf{40.99} & \textbf{72.07}  & \textbf{95.68} &  \textbf{89.74} \\

\bottomrule
\end{tabularx}
  \label{table:isolated-slr}
\end{table}

\begin{table*}[t]
  \caption{{Fine tuning in CSLR datasets. Results are reported in WER}}

\centering

\begin{tabular}{l *{6}{c}}
\toprule

  & \multicolumn{6}{c}{\textbf{Datasets}}\\
   \cmidrule(l){2-7}
   \textbf{Method} & Phoenix SD & Phoenix SI & CSL SI  & CSL SD & GSL SI & GSL SD   \\
    \midrule
    \centering
     & {Val. / Test}  & Val. / Test & Test & Test & Test & Test  \\
I3D+BLSTM (Kinetics) & 53.81 / 51.27& 65.53 / 62.38 & 23.19  & 72.39 & 34.52 & 75.42 \\
I3D+BLSTM (Kinetics + ASL 1000) & 40.89 / \textbf{40.49} & 59.60 / \textbf{58.36}     &  \textbf{16.73} &  \textbf{64.72} & \textbf{27.09} & \textbf{71.05}    \\
\bottomrule
\end{tabular} 

  \label{table:transfer_learning_asl}
\end{table*}

The referenced models, depending on the dataset, have been modified as follows: In SubUNets, AlexNet \cite{krizhevsky2012imagenet} is used as feature extractor instead of CaffeNet \cite{jia2014caffe}, as they share a similar architecture. Additionally, for the CSL and GSL datasets, we reduce the bidirectional LSTM hidden size by half, due to computational space complexity. In the isolated setup, the LSTM layers of SubUNets are trained along with the feature extractor. In order to achieve the maximum performance of GoogLeNet+TConvs, a manual customization of TConvs 1D CNN kernels and pooling sizes is necessary. The intuition behind it, is that the receptive field should be approximately covering the average gloss duration. Each 1D CNN layer includes 1024 filters. In CSL, the 1D CNN are set with kernel size 7, stride 1 and the max-pooling layers with kernel sizes and strides equal to 3, to cover the average gloss duration of 58 frames. For the GSL dataset the TConvs are tuned with kernel sizes equal to 5 and pooling sizes equal to 3. 
In order to deploy 3D-ResNet+BLSTM and I3D+BLSTM in a CSLR setup, a sliding window technique is adopted in the input sequence. Window size is set to cover at least the average gloss duration, while the stride to ensure a minimum 50\% overlap. Then, a 2-layer bidirectional LSTM is added to model the long-term temporal correlations in the feature sequence. In CSL, the window size is set to 50 and stride 36, whereas in GSL the window size is set to 25, with stride equal to 12. I3D+BLSTM and 3D-ResNet+BLSTM are initialized with weights pretrained on Kinetics. Also, for the 3D-ResNet+BLSTM method, we omit the attentional decoder from the original paper, keeping the 3D-ResNet+BLSTM model. In all CSLR experiments, proximal transfer learning is selected for model pretraining. 
In initial experiments, it was observed that by training with StimCTC, all baseline models were unable to converge. The main reason is that the networks produce unstable output probability distributions in the early stage of training. On the contrary, introducing $L_{stim}$ in the late training stage, constantly improved the networks' performance. The overall best results were obtained with EnStimCTC. The reason is that, while the entropy term $H$ introduces more variability in the early optimization process, convergence is hindered on the late training stage. By removing $H$ and introducing $L_{stim}$, the possible alignments generated by EnCTC are filtered. Regarding the hyper-parameters of the selected criteria, a tuning was necessary. For EnCTC, the hyperparameter $\phi$ is varied in the range of $0.1$ and $0.2$. For EnStimCTC, $\lambda$ is set to 1. Concerning $\theta$, evaluations for $\theta=0.1, 0.2, 0.5, 1$ are performed. The best results were obtained with $\theta=0.5$ and $\phi=0.1$.

\begin{table}[b]
    \caption{{Comparison of pretraining schemes: results of the I3D+BLSTM architecture, as measured in test WER, using multiple fully-supervised approaches before training in CSLR}}
    \centering
    \begin{center}
        \begin{tabular}{l *{4}{c}}
            \toprule & \multicolumn{2}{c}{\textbf{Datasets}}\\
            \cmidrule(l){2-3}
            \textbf{Method} & CSL SI & GSL SI\\
    \midrule
    \centering
      & Test &  {Val. / Test}   \\
            Transfer learning only from Kinetics & 23.19 & 33.94 / 34.52\\
            SubUNets alignments & \textbf{5.94} & 18.43 / 20.00\\
            Uniform alignments & 16.98 & 27.30 / 29.08\\
            Transfer learning from ASL-1000 & 16.73 & 25.89 / 27.09\\
            Proximal transfer learning & 6.45 & 8.78 / \textbf{8.62}\\
            \bottomrule
        \end{tabular} 
    \end{center}
    \label{table:pretrain comparison}
\end{table}
\subsection{Experimental results}
\begin{table*}[!t]
  \caption{Reported results in continuous SD SLR datasets, as measured in WER. Pretraining is performed on the respective isolated.}
  \label{table:Signer-Dependent}
  \centering
\begin{tabularx}{\textwidth}{@{}l @{\extracolsep{\fill}} *{9}{c}@{}}  
\toprule
  & \multicolumn{9}{c}{\textbf{Signer Dependent Datasets}} \\
   \cmidrule(l){2-10}
   
  & \multicolumn{3}{c}{Phoenix SD}
  & \multicolumn{3}{c}{CSL SD} 
  & \multicolumn{3}{c}{GSL SD}  \\
   \cmidrule(l){2-10}
   
 \textbf{Method} & CTC & EnCTC & EnStimCTC & CTC & EnCTC & EnStimCTC & CTC & EnCTC & EnStimCTC \\
 \midrule
 & Val. / Test & Val. / Test & Val. / Test & Test &Test & Test &Val. / Test & Val. / Test & Val. / Test\\
 
 SubUNets \cite{camgoz2017subunets}  &  30.51/\textbf{30.62}  &  32.02/\textbf{31.61}& 29.51/29.22 & 78.31 & 81.33 & 80.13 & 52.79/54.31 & 58.11/60.09 & 55.03/57.49  \\

GoogLeNet+TConvs \cite{cui2019deep}  & 32.18/31.37    &  31.66/31.74   & 28.87/\textbf{29.11} & 65.83 &  \textbf{64.04} & 64.43 & 43.54/\textbf{48.46}  & 42.69/\textbf{44.11} & 38.92/\textbf{42.33} \\

3D-ResNet+BLSTM \cite{pu2019iterative}  & 38.81/37.79  &  38.80/37.50   &   36.74/35.51 & 72.44 &  70.20 & 68.35 & 61.94/68.54  & 63.47/66.54 &  57.88/61.64  \\
I3D+BLSTM \cite{carreira2017quo}       & 32.88/31.92    &  32.60/32.70   &  31.16/31.48 & \textbf{64.73} & 64.06 & \textbf{60.68} & 51.74/53.48  & 51.37/53.48 & 49.89/49.99 \\
\bottomrule
\end{tabularx} 
\end{table*}
\begin{table*}[!t]
  \caption{Reported results in continuous SI SLR datasets, as measured in WER. Pretraining is performed on the respective isolated.}
  \centering
    
\begin{tabularx}{\textwidth}{@{}l @{\extracolsep{\fill}} *{9}{c}@{}}

\toprule
  & \multicolumn{9}{c}{\textbf{Signer Independent Datasets}}\\
   \cmidrule(l){2-10}
  & \multicolumn{3}{c}{Phoenix SI} &  \multicolumn{3}{c}{CSL SI}  & \multicolumn{3}{c}{GSL SI}    \\
  \cmidrule(l){2-10}
 \textbf{Method} & CTC & EnCTC & EnStimCTC & CTC & EnCTC & EnStimCTC & CTC & EnCTC & EnStimCTC \\
 \midrule
  & Val. / Test & Val. / Test & Val. / Test & Test &Test & Test & Val. / Test & Val. / Test & Val. / Test\\
SubUNets \cite{camgoz2017subunets}   & 56.56/55.06   &  55.59/53.42   &  55.01/54.11 & \textbf{3.29} & 5.13 &  4.14    & 24.64/24.03 & 21.73/20.58 & 21.65/20.62 \\

GoogLeNet+TConvs \cite{cui2019deep} & 46.70/\textbf{46.67}   & 47.14/\textbf{46.70}  & 46.42/\textbf{46.41} & 4.06  &\textbf{2.46} &  \textbf{2.41}  &  8.08/\textbf{7.95} & 7.63/6.91 &6.99/6.75 \\
{3D-ResNet+BLSTM} \cite{pu2019iterative}  &  55.88/53.77   &  54.69/54.57   &  52.88/50.98 & 19.09  & 13.36 & 14.31   & 33.61/33.07 & 27.80/26.75 & 25.58/24.01
\\
{I3D+BLSTM} \cite{carreira2017quo} & 55.24/54.43   &  54.42/53.92   &  53.70/52.71  & 6.45  & 4.26 &  2.72 & 8.78/8.62 & 7.69/\textbf{6.55} & 6.63/\textbf{6.10}   \\
\bottomrule
\end{tabularx} 
  \label{table:cont-slr-si}
\end{table*}

\subsubsection{ Evaluation on ISLR datasets}

In Table \ref{table:isolated-slr}, quantitative results are reported for the isolated setup. Classification accuracy is reported in percentage. It can be seen that 3D baseline methods achieve higher gloss recognition rate than 2D ones. I3D+BLSTM clearly outperforms other architectures in this setup, by a minimum margin of 2.2\% to a maximum of 21.6\%. I3D+BLSTM and 3D-ResNet+BLSTM were pretrained on Kinetics, which explains their superiority in performance as they contain motion priors. The 3D CNN models achieve satisfactory results in datasets created under laboratory conditions, yet in challenging scenarios, I3D+BLSTM clearly outperforms 3D-ResNet+BLSTM. Specifically in ASL 1000, where glosses are not executed in a controlled environment, only I3D+BLSTM is able to converge. SubUNets performed poorly or did not converge at all and its results are deliberately excluded. SubUNets' inability to converge, may be due to their large number of parameters (roughly 125M). From these experiments, it is concluded that 3D CNN approaches have better short-term temporal modeling capabilities than 2D CNNs, even when the latter are equipped with temporal modeling modules such as TConvs. 

\subsubsection{Comparison of pretraining schemes}
In Table \ref{table:transfer_learning_asl}, {I3D+BLSTM} is fine-tuned on CSLR datasets with CTC in 2 configurations: a) using the pretrained weights from Kinetics, and b) further pretraining the Kinetics initialized network in ASL 1000. In this manner, it is further evaluated how transfer learning from a big, diverge dataset impacts model performance in CSLR. Results are significantly improved with further pretraining in ASL 1000, which is expected due to the task relevance.

Table \ref{table:pretrain comparison} presents an evaluation of the impact of multiple pretraining schemes. The following cases are considered:

\begin{itemize}
    \item direct transfer learning from a large-scale human action recognition dataset (Kinetics) without any additional SLR-specific pretraining,
    \item directly train a shallow model (i.e., SubUNets) without pretraining, to obtain initial pseudo-alignments,
    \item assume uniform pseudo-alignments over input video for each gloss in a sentence,
    \item transfer learning from a large-scale isolated dataset (ASL-1000) and
    \item proximal transfer learning from the respected datasets isolated.
\end{itemize}

Experiments are conducted on the CSL SI and the GSL SI evaluation sets, since they have annotated isolated subsets for proximal transfer learning. For the particular experiment I3D+BLSTM is used, since it is the best performing model in isolated setup (Table \ref{table:isolated-slr}). SubUNets are chosen to infer the initial pseudo-alignments, because pretraining is not required by design. Training was performed with the traditional CTC criterion. 

In CSL SI, the best strategy seems to be pretraining on pseudo-alignments and proximal transfer learning, achieving WERs of 5.94\% and 6.54\%, respectively. On the contrary, on GSL SI the best results are achieved with proximal transfer learning with a test WER of 8.62\% compared to pseudo-alignments that a WER of 20.00\% is achieved. In general, producing pseudo-alignments requires more training time, while in some cases it might be noisy, resulting in inaccurate gloss boundaries i.e, on the CSL dataset. Therefore, pretraining with a proximal isolated subset is preferred over pretraining with pseudo-alignments.

\subsubsection{Evaluation on CSLR datasets}

In Tables \ref{table:Signer-Dependent} and \ref{table:cont-slr-si}, quantitative results regarding CSLR are reported. The selected architectures are evaluated on CSLR datasets in both SD and SI subsets, using the proposed criteria. Training with EnCTC, needs more epochs to converge, due to the fact that a greater number of possible paths is explored, yet it converges to a better local optimal. Overall, EnCTC shows an average improvement of 1.59\% in WER. A further reduction of 1.60\% in WER is observed by adding StimCTC. 
It can be seen that the proposed EnStimCTC criterion improves recognition in all datasets by an overall WER gain of 3.26\%. In the reported average gains SubUNets are excluded due to performance deterioration.

\begin{figure}[t]
 \centering
 \includegraphics[width=\linewidth]{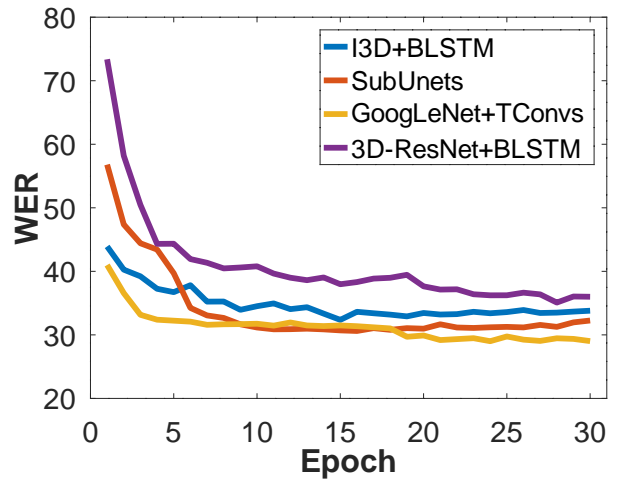}
  \caption{Validation WER of the implemented architectures on Phoenix SD dataset trained with EnStimCTC loss.}
  \label{fig:nn_phoenix_comparison}
\end{figure}

\begin{table}[b]
    \caption{{Reported results using depth modality measured in test WER}}
    \centering
    \begin{center}
        \begin{tabular}{l *{4}{c}}
            \toprule & \multicolumn{2}{c}{\textbf{Datasets}}\\
            \cmidrule(l){2-3}
            \textbf{Method} & GSL SI & GSL SD\\
    \midrule
    \centering
      & {Val. / Test}  &  {Val. / Test}   \\
      SubUNets \cite{camgoz2017subunets} & 52.07/50.45 & 73.11/ 72.76\\
GoogLeNet+TConvs \cite{cui2019deep} &49.08/44.70 & 85.90/84.31\\
      3D-ResNet+BLSTM \cite{pu2019iterative} & 53.12/51.68 & 75.04/74.65 \\
{I3D+BLSTM} \cite{carreira2017quo} & \textbf{31.96}/\textbf{29.03} & \textbf{65.50}/\textbf{64.98}\\
            \bottomrule
        \end{tabular} 
    \end{center}
    \label{table:depth}
\end{table}


On the Phoenix SD subset, all models benefit from training with EnStimCTC loss and have reduced WER. SubUNets have a WER of 29.51\% on the validation set and 29.22\% on the test set, which is an average reduction of 12.59\% WER, compared to the original paper's results (42.1\% vs 30.62\%) \cite{camgoz2017subunets}. Furthermore, 2D-based CNNs produce similar results with negligible difference in performance. Nonetheless, our implementation of GoogLeNet+TConvs trained with EnStimCTC loss achieves a WER of 29.11\% on the test set  performing inferior to the original work. This is due to the iterative optimization that is employed in \cite{cui2019deep}. The latter currently holds the state-of-the-art performance on Phoenix SD, which is 24.43\% WER on the test set. \figurename \ref{fig:nn_phoenix_comparison}, depicts the models' WER on Phoenix SD validation set.

Similarly, on Phoenix SI, GoogLeNet+TConvs trained with EnStimCTC is the best performing setup with 46.41 \% test WER. Finally, all architectures on Phoenix SI have worse recognition performances compared to their SD, due to a reduction of more than 20\% on the training data.

\begin{figure}[t]
 \centering
 \includegraphics[width=\linewidth]{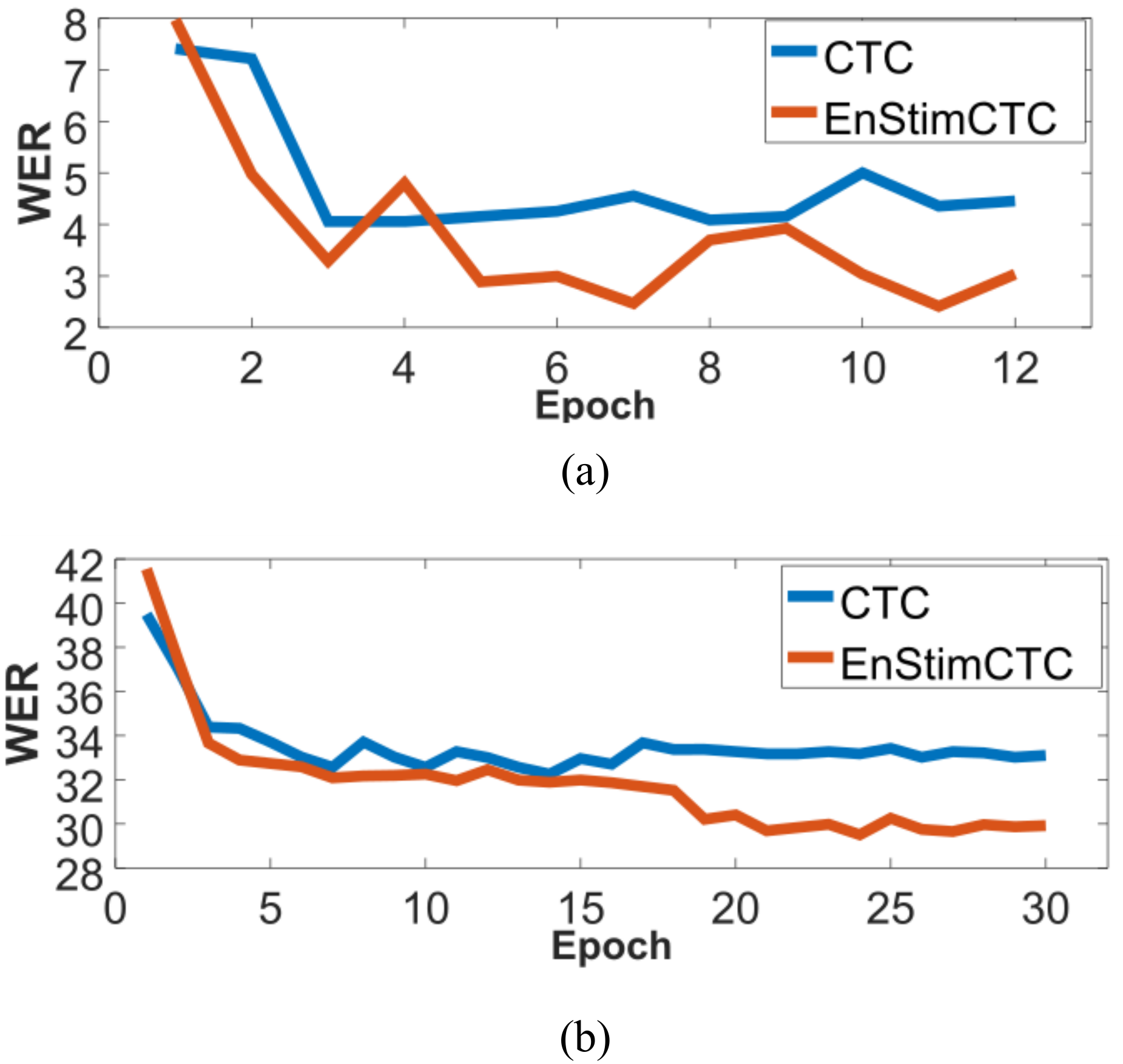}
  \caption{Comparison of validation WER of CTC and EnStimCTC criteria with GoogLeNet+TConvs on a) CSL SI and b) Phoenix SD datasets.}
  \label{fig:cui_enstim_vs_ctc}
\end{figure}

\begin{figure*}[t]
 \centering
 \includegraphics[width=0.99\textwidth]{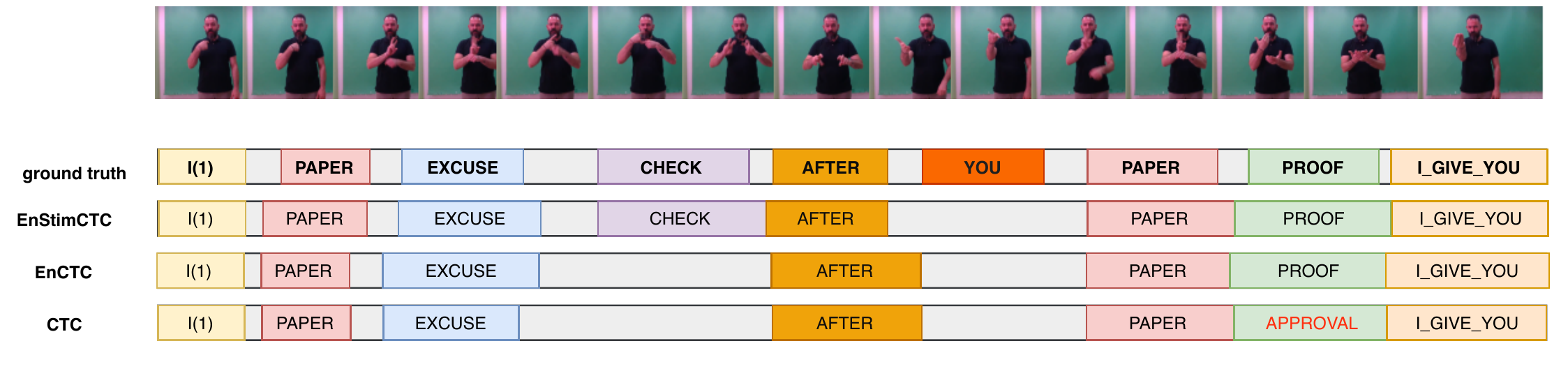}
  \caption{Visual comparison of ground truth alignments with the predictions of the proposed training criteria. GoogLeNet+TConvs is used for evaluation on the GSL SD dataset.}
  \label{fig:algnments_ctc}
\end{figure*}

\begin{figure*}[t]
 \centering
 \includegraphics[width=0.99\textwidth]{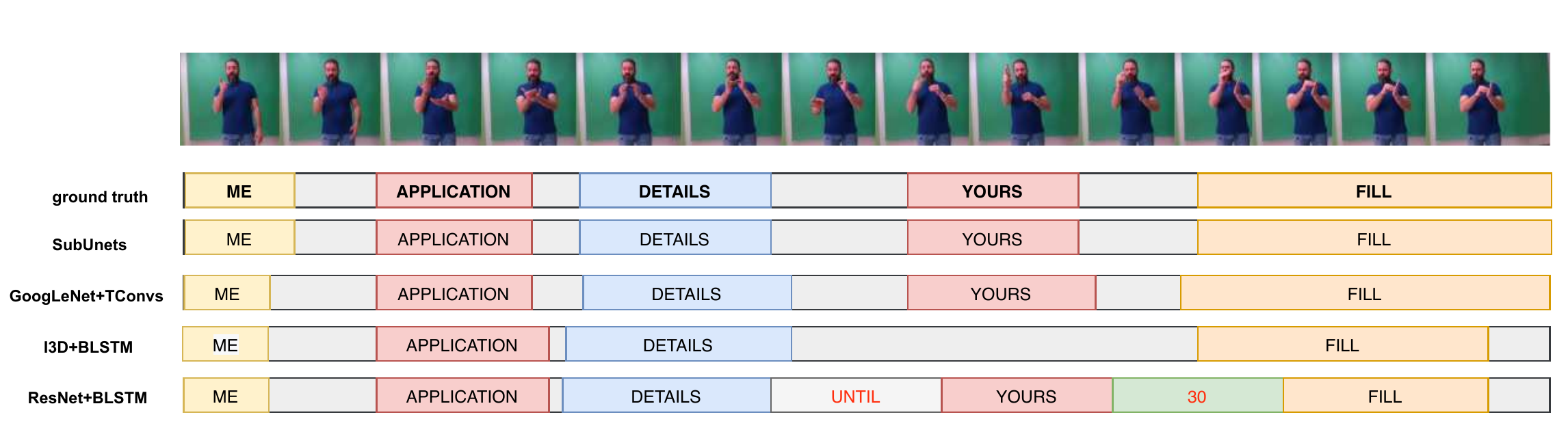}
  \caption{Visual comparison of ground truth alignments with the predictions of each method. The methods are trained with EnStimCTC loss and are evaluated on the GSL SΙ dataset.}
  \label{fig:algnments_models}
\end{figure*}

{On the CSL SI dataset all methods, except for 3D-ResNet+BLSTM, have comparable recognition performance. They achieve high recognition accuracy due to the large size of the dataset and the small size of the vocabulary. I3D+BLSTM seems to benefit the most when trained with EnStimCTC, with 3.73\% absolute WER reduction. GoogLeNet+TConvs has the best performance with 2.41\% WER, with an absolute reduction is 1.65\% less compared to CTC training} (\figurename{} \ref{fig:cui_enstim_vs_ctc}). This method outperforms the current state-of-the-art method on CSL SI \cite{yang2019sf} {by an absolute WER improvement of 1.39\%. On CSL SD, all models perform considerably worse compared to CSL SI, which is a challenging CSLR task since the dataset has a small combination of unique sentences (100 sentences) and the test set contains different sentences, i.e., unseen sentences, from those on the training set, with 6\% of the sentences on CSL SD used for testing. I3D+BLSTM trained with EnStimCTC loss has the best performance with a WER of 60.68\% on the test set.}



On GSL SI, I3D+BLSTM and GoogLeNet+TConvs trained with EnStimCTC have comparable performance, with 6.63\%/6.10\% and 6.99\%/6.75\% WER in validation and test set, respectively. On  GSL SD, GoogLeNet+TConvs has the best recognition rate with 38.92\%/42.33\% in validation and test set, respectively, yet is still worse than its results on GSL SI. 
The reported performances on the GSL dataset are justified because it has only with 331 unique sentences. As a result, its SD evaluation and test sets contain approximately 18\% unseen sentences (60 sentences). For this reason, all models tend to predict combination of glosses similar to the ones seen during training, and as a result the performances are inferior compared to the SI split.

\subsubsection{Evaluation on GSL using depth modality}
In Table \ref{table:depth}, experiments using the depth modality on GSL SI and SD datasets are reported. I3D+BLSTM has the best performance with WER of $31.96/29.03 \%$ on validation and test sets of GSL SI, respectively. On the GSL SD subset, all models have a decreased performance compared to GSL SI, with the best WER being $65.50/64.98 \%$ achieved by I3D+BLSTM. It should be noted that the depth counterpart of the GSL is not as expressive as the RGB modality, since details about finger movements and facial expressions are not visible in the depth videos.

\section{Discussion}  
\label{sec:Discussion}

\subsection{Performance comparison of implemented architectures}
From the group of experiments in isolated SLR in Table \ref{table:isolated-slr}, it was experimentally shown that 3D-CNN methods are more suitable for isolated gloss classification compared to the 2D-CNN models, using only RGB image data. This is justified by the fact that 2D CNNs do not model dependencies between neighbouring frames, where motion features play a crucial role in classifying a gloss. Furthermore, 3D CNNs may contain motion priors as they are pretrained on Kinetics. Overall, I3D is the best-performing method, as it captures better the intra-gloss dependencies.

The advantage of 3D inception layer lies in the ability to project multi-channel spatio-temporal features in dense, lower dimensional embeddings. This leads in accumulating higher semantic representations (more abstract output features). In the opposite direction, the 3D-ResNet+BLSTM lacks the modeling capabilities of I3D, since it has a smaller feature dimension (512 vs 1024) and the 3D-ResNet-18 backbone which is employed in the 3D-ResNet+BLSTM, is inferior to I3D in the Kinetics dataset  (54.2 Top-1 score VS 68.4) \cite{hara2018can}. Thus, in an architectural level one can state that the inception modules of I3D are more effective than residual blocks with identity skip connections for isolated SLR.

Modeling intermediate short temporal dependencies was experimentally shown (Tables \ref{table:Signer-Dependent}, \ref{table:cont-slr-si}) to enhance the CSLR performance. The implemented 3D CNN architectures directly capture spatio-temporal correlations as intermediate representations. The design choice of providing the input video in a sliding window restricts the temporal receptive field of the network. Based on a sequential structure, architectures such as GoogLeNet+TConvs achieve the same goal, by grouping consecutive spatial features. Such a sequential approach can be proved beneficial in many datasets, given that spatial filters are well-trained. For this reason, such approaches require heavy pretraining in the backbone network. The superiority in performance of the implemented sequential approach is justified in the careful manual tuning of temporal kernels and strides. However, manual design significantly downgrades the advantages of transfer learning. The sliding window technique, can be easily adapted based on the particularities of each dataset without changing the structure of the network (i.e., kernel sizes or strides), making 3D CNNs more scalable and suitable for real-life applications. To summarize, both techniques aim to approximate the average gloss duration. This is interpreted as a guidance in models, based on the statistics of the SL dataset. On the other hand, utilizing only LSTMs to capture the temporal dependencies (i.e., SubUNets), results in an ineffective modeling of intra-gloss correlations. LSTMs are designed to model the long-term dependencies that correspond to the inter-gloss dependencies. Taking a closer look at the predicted alignments of each approach in \figurename{} \ref{fig:algnments_models}, it is noticed that 3D architectures do not provide as precise gloss boundaries for true positives as the 2D ones. We strongly believe that this is the reason that 3D models benefit more from the introduced variations of the traditional CTC.

\subsection{Comparison between CTC variations}
The reported experimental results exhibit the negative influence of CTC's drawbacks (overconfident paths and conditional independence assumption) in CSLR. EnCTC's contribution to alleviate the overconfident paths, is illustrated in \figurename{} \ref{fig:algnments_ctc}. The ground truth gloss ``PROOF" is recognized with the introduction of $H$, instead of ``APPROVAL". The latter has six times higher occurrence frequency. After a careful examination of the aforementioned signs, one can notice that these signs are close in terms of hand position and execution speed, which justifies the depicted predictions. Furthermore, it is observed that EnCTC boosts performance mostly on CSL SI and GSL SI, due to the limited diversity and vocabulary. It can be highlighted that EnCTC did not boost SubUNets' performance. The latter generates per frame predictions ($T=N$), wherein the rest approaches generate grouped predictions ($T \approx \frac{N}{4}$). This results in a significantly larger space of possible alignments that is harder to be explored from this criterion. From \figurename{} \ref{fig:algnments_ctc}, it can be visually validated that EnStimCTC remedies the conditional independence assumption. For instance, the gloss ``CHECK" was only recognised with stimulated training. By bringing closer predictions that correspond to the same target gloss, the intra-gloss dependencies are effectively modeled. In parallel, the network was also able to correctly classify transitions between glosses as $blank$. It should be also noted that EnStimCTC does not increase time and space complexity during inference.

\subsection{Evaluation of pretraining schemes} 
Due to the limited contribution of CTC gradients in the feature extractor, an effective pretraining is mandatory. As shown in \figurename{} \ref{fig:nn_phoenix_comparison}, pretraining significantly affects the starting WER of each model. Without pretraining, all models congregate around the most dominant glosses, which significantly slows down the CSLR training process and limits the learning capacity of the network. Fully supervised pretraining is interpreted as a domain shift to the distribution of the SL dataset that speeds up the early training stage in CSLR.
Regarding the pretraining scheme, in datasets with limited vocabulary and gloss sequences (i.e., CSL), inferring initial pseudo-alignments proved slightly beneficial, as shown in Table \ref{table:pretrain comparison}. This is explained due to the fact that the data distribution of the isolated subset had different particularities, such as sign execution speed. However, producing initial pseudo-alignments is time consuming. Hence, the small deterioration in performance is an acceptable trade-off between recognition rate and time to train. 

\begin{figure}[t]
 \centering
 \includegraphics[width=\linewidth]{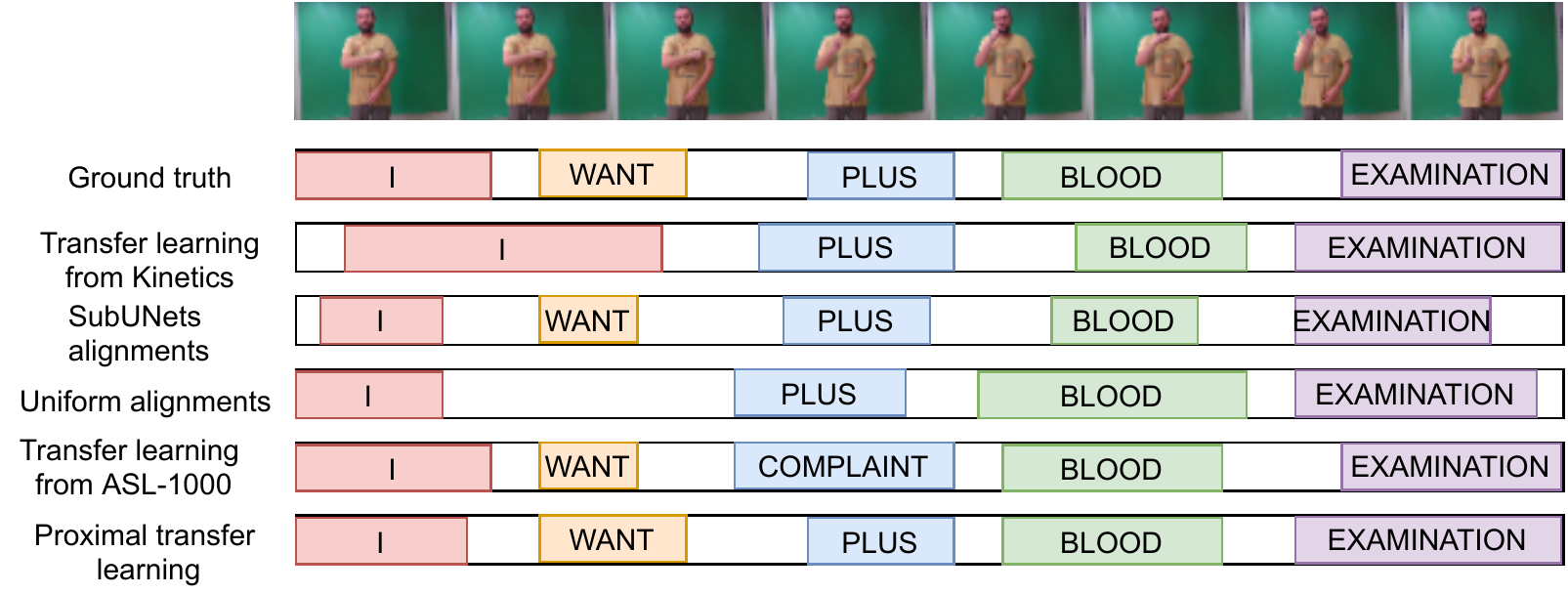}
  \caption{Visual comparison of the alignments produced by I3D+BLSTM with different pretraining schemes on the GSL SΙ dataset.}
  \label{fig:table4_fig}
\end{figure}

The proposed GSL dataset contains nearly double the vocabulary and roughly three times the number of unique gloss sentences, with less training instances. More importantly, in GSL the isolated subset draws instances from the same distribution as the continuous one. \figurename{} \ref{fig:table4_fig} illustrates the alignments produced by I3D+BLSTM given different pretraining schemes. It can be stated that proximal transfer learning significantly outperforms training with pseudo-alignments in this setup, both quantitatively and qualitatively. Thus, by annotating the same videos one can create larger SL datasets and more efficient CSLR systems for new SL datasets.

\section{Conclusions and future work} \label{sec:conclusion}
In this paper, an in-depth analysis of the most characteristic DNN-based SLR model architectures was conducted. Through extensive experiments in three publicly available datasets, a comparative evaluation of the most representative SLR architectures was presented. Alongside with this evaluation, a new publicly available large-scale RGB+D dataset was introduced for the Greek SL, suitable for SLR benchmarking. Two CTC variations known from other application fields, EnCTC \& StimCTC, were evaluated for CSLR and it was noticed that their combination tackled two important issues, the ambiguous boundaries of adjacent glosses and intra-gloss dependencies. Moreover, a pretraining scheme was provided, in which transfer learning from a proximal isolated dataset can be a good initialization for CSLR training. The main finding of this work was that while 3D CNN-based architectures were more effective in isolated SLR, 2D CNN-based models with an intermediate per gloss representation achieved superior results in the majority of the CSLR datasets. In particular, our implementation of GoogLeNet+TConvs, with the proposed pretraining scheme and EnStimCTC criterion, yielded state-of-the-art results on CSL SI.

Concerning future work, efficient ways for integrating depth information that will guide the feature extraction training phase can be devised. Moreover, another promising direction is to investigate the incorporation of more sequence learning modules, like attention-based approaches \cite{camgoz2020sign}, in order to adequately model inter-gloss dependencies. Future SLR architectures may be enhanced by fusing highly semantic representations that correspond to the manual and non-manual features of SL, similar to humans. Finally, it would be of great importance for the Deaf-non Deaf communication to bridge the gap between SLR and SL translation. Advancements in this domain will drive research to SL translation as well as SL to SL translation, which have not yet been thoroughly studied.


\section{Acknowledgements}
This work was supported by the Greek General Secretariat of Research and Technology under contract \Tau1\Epsilon$\Delta$\Kappa-02469 EPIKOINONO. The authors would like to express their gratitude to Vasileios Angelidis, Chrysoula Kyrlou and Georgios Gkintikas from the Greek sign language center\footnote[4]{\url{https://www.keng.gr/}} for their valuable feedback and contribution to the Greek sign language capturings.

\bibliographystyle{IEEEtran}
\bibliography{references}
\end{document}